\documentclass[runningheads]{llncs}
\usepackage[T1]{fontenc}
\usepackage{graphicx}
\usepackage{amsmath}
\usepackage{multirow}
\usepackage{xcolor}
\usepackage{colortbl}
\usepackage{pifont}
\usepackage{booktabs}
\usepackage{comment}

\begin{document}
\title{Detail-Enhanced Intra- and Inter-modal Interaction for Audio-Visual Emotion Recognition}
\titlerunning{Intra- \& Inter-modal Interaction for Audio-Visual Emotion Recognition}
\author{Tong Shi\orcidID{0000-0001-5913-4095} \and
Xuri Ge\orcidID{0000-0002-3925-4951} \and\\
Joemon M. Jose\orcidID{0000-0001-9228-1759}\and
Nicolas Pugeault\orcidID{0000-0002-3455-6280}\and%
Paul Henderson\orcidID{0000-0002-5198-7445}}
\authorrunning{T.~Shi \textit{et al.}}
\institute{School of Computing Science, University of Glasgow}
\maketitle             

\begin{abstract}
Capturing complex temporal relationships between video and audio modalities is vital for Audio-Visual Emotion Recognition (AVER). However, existing methods lack attention to local details, such as facial state changes between video frames, which can reduce the discriminability of features and thus lower recognition accuracy. In this paper, we propose a Detail-Enhanced Intra- and Inter-modal Interaction network (DE-III) for AVER, incorporating several novel aspects. We introduce optical flow information to enrich video representations with texture details that better capture facial state changes. A fusion module integrates the optical flow estimation with the corresponding video frames to enhance the representation of facial texture variations. We also design attentive intra- and inter-modal feature enhancement modules to further improve the richness and discriminability of video and audio representations. A detailed quantitative evaluation shows that our proposed model outperforms all existing methods on three benchmark datasets for both concrete and continuous emotion recognition. To encourage further research and ensure replicability, we will release our full code upon acceptance.

\end{abstract}

\section{Introduction}

Emotion perception is attracting ever-increasing research attention due to its wide range of applications, such as affective computing~\cite{savchenko2021facial}, human-computer interaction~\cite{busso2016msp}, and social robotics~\cite{spezialetti20robots}. Multi-modal emotion recognition, especially integrating audio and video (i.e.~AVER), is particularly important since it makes use of the information present in two modalities that are vital to human communication. Unlike single-modal emotion recognition, multi-modal emotion recognition has access to different representations of the same emotion from different modalities. This improves feature representation capabilities and distinguishability, leading to improved recognition accuracy~\cite{tsai2019multimodal,chumachenko2022self}. 

However, there are still two challenges that are the focus of ongoing research in AVER: (i) how to enhance the representation of fine details within modalities, such as tiny details of facial motion (e.g.~due to micro-expressions), and (ii) how to better leverage inter-modal associations to fully exploit the complementary information from different modalities.
Solving both will enable learning better feature representations, and improve emotion recognition accuracy

When learning features from one modality, intra-modal temporal relationship mining \cite{zhou2019exploring,ge2024mgrr,zhang2023adaptive} and feature detail enhancement \cite{wang2021micro} are important ways to make features more discriminative. For instance, \cite{zhang2023adaptive} proposed an adaptive graph attention network to explore the relationship between frames of videos for micro-expression recognition, while \cite{wang2021micro} introduced optical flow to replace face images for micro-expression recognition based on a multi-scale feature representation. However, these methods focus on the single-modal setting, and cannot exploit information from multiple modalities. \cite{zhou2019exploring} used self-attention \cite{vaswani2017attention} within each modality to enhance their representation and then fused them by a linear-based function to classify; however this cannot fully account for the complex, nonlinear relationships between audio and video.

\begin{figure}[t]
	\centering
	\includegraphics[width=1\linewidth]{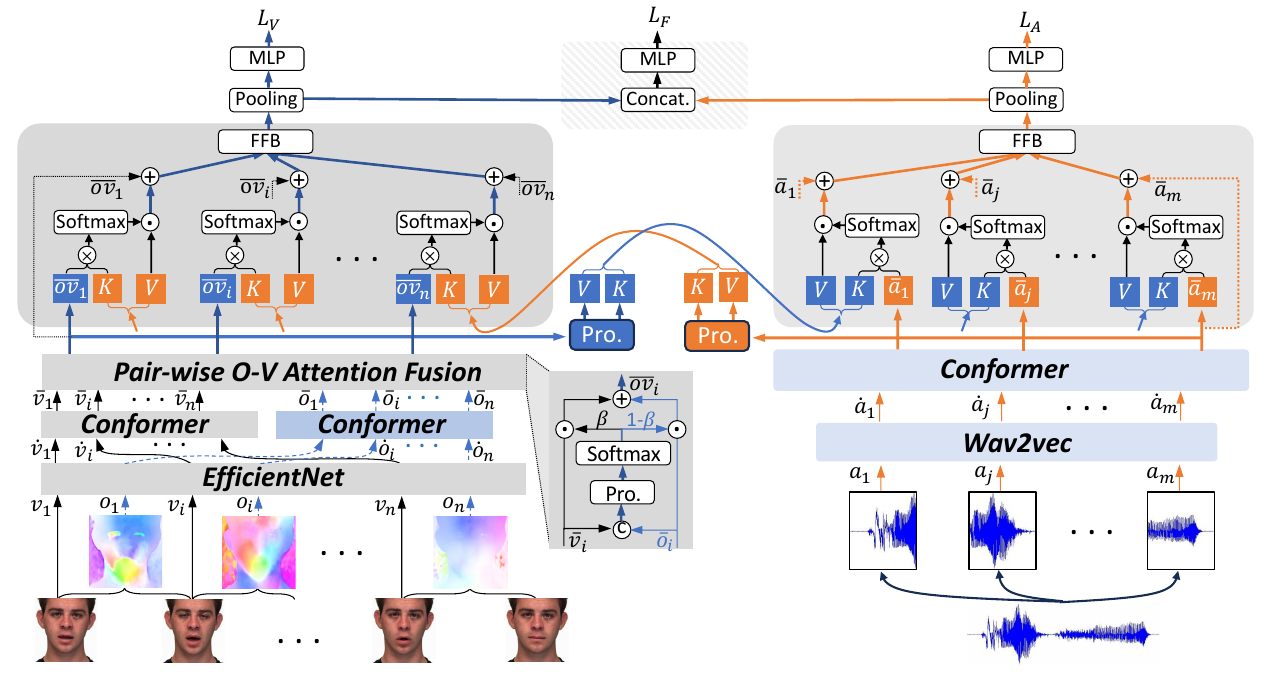}
	\vspace{-1em}
	\caption{
 Overview of our proposed method \textit{DE-III}. Given video frames $v_i$ and audio fragments $a_i$, we extract features and pass these through separate Conformer encoders. We introduce explicit information about facial motions -- captured by optical flow $o_i$ -- to enhance video feature representations, with a new pair-wise O-V attention fusion module that effectively integrates the information from optical flow and video frames. We propose an inter-modal feature enhancement module (large boxes near top) to attentively fuse the associated audio and video representations in both directions, i.e.~audio-to-video and video-to-audio. During training, the final emotion predictions are calculated independently from three sets of features: the video features albeit with audio information fused (i.e.~without the model components in the chequered box); the converse using the audio features; and finally using both sets of features after a further fusion stage. During inference, we use the prediction head that performed best on validation data.
 }
	\label{fig:frame}
\end{figure}

Multi-modal approaches have recently become mainstream \cite{ma2019end,nie2020c,yin2023msa} since considering both audio and video further improves representations, by fusing information in associated video frames and audio fragments. For example, \cite{chumachenko2022self} explored the effectiveness of different variants of transformer-based inter-modal attention mechanisms for AVER and showed inter-modal interaction can significantly improve performance.  However, although inter-modal interaction improves recognition, these methods do not investigate modeling temporal relationships within each modality. \cite{goncalves2022auxformer} adopts a multi-branch joint auxiliary training method, designing independent audio and video branches and multi-modal fusion to enhance feature relationships, which greatly improves recognition performance. \cite{gong2022uavm,goncalves2023versatile} used a network shared across modalities to encourage consistency of the multi-modal feature space. However, since different modalities have different feature distributions and properties, a shared network may not fully capture the unique characteristics of each modality, resulting in information loss.

Most of the relationship modeling strategies mentioned above~\cite{tarantino2019self,conformer2021,gong2022uavm,goncalves2023versatile} model temporal relationships based on implicit appearance representation of video frames and audio fragments, but ignore an inherent challenge of AVER -- that in video features the frame-to-frame variations of faces are much weaker than in audio. For example, there may be significant changes in content and intonation between two audio fragments, while there is little difference between video frames. It is clear that these missing explicit details, especially state changes between face frames of videos, may lead to reduced discriminability of feature representations during the relationship modeling process, thereby affecting the accuracy of AVER.

We address these issues by introducing a multi-modal interaction network (Figure~\ref{fig:frame}) that incorporates an explicit representation of visual detail changes between frames, and which can better fuse the complementary information from video and audio. Different from methods that directly model relationships between local regions of a facial sequence \cite{perveen2020facial,ge2021local,hu2021spatio,ge2023algrnet}, optical flow is a simple and effective way to represent the state changes between the facial frames. Optical flow can enhance the discriminability of visual representations by directly highlighting significant detail differences between frames, especially those texture changes that can express facial emotions \cite{wang2021micro}. To this end, we propose a novel detail-enhanced intra- and inter-modal interactions network (called DE-III) for AVER, which integrates explicit optical flow information into an end-to-end multi-modal interaction framework. In addition, two independent multi-modal interaction fusion mechanisms and multiple residual connections further alleviate the information loss problem in existing shared interaction strategies \cite{gong2022uavm,goncalves2023versatile}.
Our main contributions are as follows:
\begin{itemize} 
\item  we explicitly capture detail changes between video frames using optical flow, and integrate this information using a lightweight attentive fusion module;
\item  we design novel detail-enhanced intra- and inter-modal interaction modules for the video and audio modalities, which can effectively fuse associated information of one modality into the other modality and reduce information loss by residual connections. 
\end{itemize}
We evaluate the resulting model and several variants on three widely used benchmarks and obtain highly competitive results including a new state-of-the-art on multiple metrics, e.g.~83.7\% F1-Micro score on CREMA-D, 82.7\% accuracy on RAVDESS and the highest scores on MSP-IMPROV with 89.3\%, 88.7\% and 85.8\% for valence, arousal and dominance.

\section{Related Work}

Emotion recognition has received a significant amount of attention in the computer vision community. Numerous methods \cite{wang2020suppressing,zhang2021relative,ruan2021feature,li2022eeg} have been proposed to solve this task by using different data modalities, such as images, speech and text. These methods can be divided into two main kinds: unimodal methods (that input just one modality), and multi-modal methods (that input two or more modalities). Our proposed DE-III belongs to the latter category, combining audio and video modalities to improve the performance of emotion recognition.

\subsection{Unimodal Emotion Recognition}
Unimodal emotion recognition methods \cite{wang2020suppressing,acheampong2021transformer,zhang2021relative,ruan2021feature,li2022eeg} focus on application scenarios where only one kind of data is available; they design feature enhancement and interaction methods based on the inherent properties of the corresponding modality. The most common methods are text-based \cite{wang2020emotion,deng2021survey,acheampong2021transformer} and image-based \cite{wang2020suppressing,zhang2021relative,ruan2021feature}. For example for text, \cite{acheampong2021transformer} present a BERT-based model to explore the importance of context extraction in texts for emotion recognition. One work by \cite{shrivastava2019effective} proposed one sequence-based convolutional neural network to detect human emotion from big data.
However, it is harder to to accurately predict human emotions from a text transcription compared to using richer modalities such as images or videos. 
For image data, \cite{ruan2021feature} proposed feature decomposition and reconstruction learning for effective facial image expression recognition. \cite{mittal2020emoticon} introduced the image depth information to improve the context information of images, which improved the representation capability and thus recognition accuracy. Moving to video, \cite{ben2021video} introduced facial micro-expression analysis methods that can improve emotion recognition by capturing richer contextual sequence information than static images. Although unimodal emotion recognition has achieved substantial progress and delivers promising results, it is inherently limited by having less information available than multi-modal approaches.

\subsection{Multi-modal Emotion Recognition}
Recently, multi-modal emotion recognition has become mainstream \cite{mittal2020m3er,chudasama2022m2fnet,goncalves2022auxformer,conformer2021,goncalves2023versatile,gong2022uavm,chumachenko2022self,goncalves2023learning,tsai2019multimodal} due to its ability to fully exploit the complementary information present in different modalities. For instance, \cite{chumachenko2022self} explored the effectiveness of different variants of transformer-based inter-modal attention mechanisms for audio-video emotion recognition and showed that inter-modal interaction can significantly improve performance. \cite{maji2023multimodal} showed that combining audio and with a corresponding text transcription improves the representation ability of features, since audio captures details of intonation, while text captures semantics more explicitly. Moreover, \cite{chudasama2022m2fnet} fused three modalities (audio, text and vision), further improving recognition accuracy.
The above works indicate that combining multiple modalities can significantly enhance the discrimination ability of fused representations and thus the recognition performance. In this work, we study multimodal-based emotion recognition, specifically for audio-video emotion recognition (AVER). The most similar works to ours are \cite{gong2022uavm,goncalves2023versatile}, both of which used a transformer-based architecture that is shared across video and audio modalities to encourage consistency of the multi-modal feature space. However, their proposed shared network cannot fully capture the unique feature distributions of each modality, such as explicit facial state changes between face video frames, resulting in the loss of information during the multimodal relationship modeling process.
Unlike \cite{goncalves2022auxformer,goncalves2023learning,tsai2019multimodal}, which adopt attention-based neural network to effectively process and integrate audio modalities, our model not only learns the intra-relationships within video feature representations but also models the inter-relationships when attentively fuses the audio representation. Our proposed model augments video features with optical flow information before fusing with the audio features. Unlike traditional methods \cite{kemmou2024automatic} that directly combine the optical flow features with visual representations, we use Conformer \cite{conformer2021} networks to extract context-aware features, and design a novel pairwise O-V attention fusion module to combine them.

\section{Proposed Model}
\label{sec:method}

The overall framework of our proposed model DE-III is shown in Figure~\ref{fig:frame}. 
We first extract video and audio features, then enhance their representative power through temporal relationship modelling within their respective modalities, also fusing optical flow information with the video features to better capture detail changes. Then, the inter-modal feature enhancement module performs attention-weighted fusion of each modality's information with the other modality.

\subsection{Audio Self-enhancement Module}
To represent the information in audio, we use a pre-trained wav2vec model \cite{hsu2021robust} to embed the extracted audio fragments. Specifically, we split a given audio clip into a sequence of $m$ fragments $A = \{a_1,\, a_2,\, \ldots,\, a_m\}$ using a sliding window. Then we use the wav2vec-large-robust model to extract corresponding fragment-level representations $\dot{A} = \{\dot{a}_1, \dot{a}_2,\, \ldots,\, \dot{a}_m\}$. Next, a Conformer encoder \cite{conformer2021} (a transformer-based model with convolutions to improve temporally-local information processing) is used to obtain enhanced audio-fragment representations $\bar{A} = \{\bar{a}_1,\, \bar{a}_2,\, \ldots,\, \bar{a}_m\}$ that account for (intra-modal) local and global temporal relationships.

\subsection{Video Pairwise Attention Enhancement Module}
Different from the audio features where contextual semantics are clear, i.e.~there is clear semantic content and significant intonation changes, in the video, subtle yet important changes in facial texture tend to be lost during feature extraction. We therefore use a pre-trained optical flow model \cite{jaegle2021perceiver} to extract the flow $o_i$ between adjacent pairs of video frames $\{v_{i-1},\, v_i\}$, where $i \in \{1,\,\ldots,\, n\}$ and $n$ is the number of video frames; this can explicitly represent fine-grained changes of facial texture such as micro-expressions. Then, we employ the widely-used EfficientNet-B2 model~\cite{savchenko2021facial}, which has been fine-tuned on VGGface2~\cite{Cao18} dataset, to extract representations for video frames and their corresponding optical flow maps; we denote these features by $\dot{V} = \{\dot{v_1},\, \dot{v_2},\, \ldots,\,\dot{v_n}\}$ and $\dot{O} = \{\dot{o_1},\, \dot{o_2},\, \ldots,\,\dot{o_n}\}$ respectively. To further enhance the representational ability of these visual features, we use two independent Conformer encoders \cite{conformer2021} to embed them into the same dimensional space as the audio modality. This also allows for subsequent inter-modal interaction. We next propose a simple and efficient pairwise O-V attention fusion module to combine the features of frames and optical flow into a joint embedding space. Specifically, we use a fully-connected (FC) layer to map the features at each time-point to two channels, then apply a softmax function \cite{chorowski2015attention} and interpret these values as weights for the frame and flow features respectively. We finally obtain the detailed-enhanced video representation $\overline{ov}_i$ by a weighted sum of linearly-projected frame features and corresponding flow features. 
Thus, we set
\begin{align}
     [\bar{o}_i: \bar{v}_i] &= [\mathrm{Conformer}(\dot{o}_i) : \mathrm{Conformer}( \dot{v}_i)], \\
     (\beta_o, \beta_v) &= \mathrm{softmax}(FC([\bar{o}_i: \bar{v}_i])), \\
     \overline{ov}_i &=  \beta_o W_o \bar{o}_i + \beta_v W_v \bar{o}_i,
\end{align}
where $[\,:\,]$ denotes concatenation along the channel dimension, $W_o$ and  $W_v$ are the linear projection parameters, and $\beta_o+\beta_v =1$. We refer to the two conformers followed by the OV-fusion as the pair-wise attention enhancement (PAE) module.

\subsection{Inter-modal Feature Enhancement Module}

Inspired by the attention mechanisms \cite{yin2023msa,goncalves2022auxformer}, we next design an inter-modal feature enhancement module (IFE) that allows each modality to attend to the other and integrate relevant information. For simplicity we describe only the audio-to-video fusion (IFE-Video); however a similar approach is used for video-to-audio. We want to allow the enhanced video frame features $\overline{ov}_i$ to attend to features of relevant audio fragments $\bar{A} = \{\bar{a}_1, \bar{a}_2, \ldots, \bar{a}_m\}$. Different from traditional self-attention \cite{vaswani2017attention} and cross-attention \cite{tsai2019multimodal}, we take the target video frame $\overline{ov}_i$ as the query to calculate the attention weights, with the audio fragment defining the keys and values after the linear projections.  Attentive fusion from another modality allows relevant modality information to be extracted and integrated, thereby improving the distinguishability of target modality representation. 
Finally, we obtain the video representations $\ddot{OV} =\{\ddot{ov_i}\}$ after IFE by adding a residual connection, and passing through a feed-forward block (FFB) which contains two linear layers. In summary, we set
\begin{align}
    s_{ij} &=  \frac{(W_{ov} \overline{ov}_i) (W_a \bar{a}_j)^T}{||W_{ov} \overline{ov}_i||\ ||W_a \bar{a}_j||} \;\; \forall \; i\in \{1,\,\ldots,\, n\},\, j \in \{1,\,\ldots,\, m\} \\
     \alpha_{ij} &= \frac{\exp(s_{ij})}{\sum_{j=1}^{m} \exp(s_{ij})}   \\
    \ddot{ov}_i &= \sum\nolimits_{j=1}^{m} \alpha_{ij} \bar{W}_a \bar{a}_j + \overline{ov}_i, 
\end{align}
where $W_{ov}$, $W_a$ and $\bar{W}_a$ are linear projection parameters. Similarly, we obtain the attention-aware video fragment representations of each audio fragment and combine them with an audio residual operation to give the final audio representations $\ddot{A} = \{ \ddot{a}_1, \ddot{a}_2, ...,\ddot{a}_m\}$.

\subsection{Feature Aggregation and Objective Function}
Since we want to make a single prediction for an entire video, we max-pool the features along the temporal axis, yielding a video-centric feature vector $\ddot{ov}^*$ from $\ddot{OV}$, and audio-centric feature vector $\ddot{a}^*$ from $\ddot{A}$ (note that $\ddot{ov}^*$ still incorporates information fused from the audio modality as described in Section~3.3, and vice-versa). We use three independent emotion prediction heads (each a multi-layer perceptron) with corresponding losses to jointly optimize different branches the model -- the video-cross loss $L_V$ (using $\ddot{ov}^*$ as input to the MLP), audio-cross loss $L_A$ (using $\ddot{a}^*$) and audio-visual fusion loss $L_F$ (using $\ddot{ov}^*$ concatenated with $\ddot{a}^*$). The overall objective function is the sum of the three losses. We use multi-class cross-entropy for datasets with discrete emotion class labels, and concordance correlation coefficient (CCC) for datasets with continuous labels. Specifically, CCC is given by
\begin{equation}
    \mathcal{L}_\mathrm{CCC}
 = 1 - \frac{2\rho \sigma_x \sigma_y}{\sigma_x^2 + \sigma_y^2 + (\mu_x - \mu_y)^2}
\end{equation} 
where $\mu_x$ and $\mu_y$ are the mean of the predicted result $\hat{y}$ and the label $y$, respectively,  $\sigma_x$ and $\sigma_y$ are their standard deviations, and $\rho$ is their Pearson correlation coefficient (a $\rho$ value close to $\pm1$ suggests a strong linear relationship, while a value of 0 signifies the absence of any linear correlation). During inference we can use predictions from any of the three heads; for our main experiments we use the prediction head that performed best on the validation data. 

\section{Experiments}

\begin{table}[t]
\centering
\setlength\tabcolsep{2pt}
\fontsize{9.5}{12}\selectfont
\caption{Comparisons with state-of-the-art methods for AVER on CREMA-D, MSP-IMPROV and RAVDESS (in \%). The best results are bold and second-best underlined.}
\label{tab:comparison}
\vspace{6pt}
\begin{tabular}{@{}ccccccccccccccc@{}}
\toprule
\multirow{2}{*}{Method} & \multicolumn{8}{c}{CREMA-D}& \multicolumn{3}{c}{MSP-IMPROV} & \multicolumn{3}{c@{}}{RAVDESS} \\
\cmidrule(){2-9}\cmidrule(lr){10-12}  \cmidrule(){13-15}  
 & \multicolumn{4}{c}{F1-Macro}& \multicolumn{4}{c}{F1-Micro}& Val.& Aro.&Dom.  & \multicolumn{3}{c}{Acc.}\\
 \midrule
Multi. \cite{tsai2019multimodal}&       \multicolumn{4}{c}{64.4}&         \multicolumn{4}{c}{69.2}&          \underline{77.5}&          \underline{76.1}&           77.8  & \multicolumn{3}{c}{78.5}\\
MMER \cite{chumachenko2022self}& \multicolumn{4}{c}{--}& \multicolumn{4}{c}{--}& --& --& -- & \multicolumn{3}{c}{\underline{81.6}}\\
UAVM \cite{gong2022uavm}  &       \multicolumn{4}{c}{74.9}& \multicolumn{4}{c}{76.9}& 47.1&  54.4&  68.7 & \multicolumn{3}{c}{--}\\ 
AuxFormer \cite{goncalves2022auxformer}    & \multicolumn{4}{c}{69.8}&  \multicolumn{4}{c}{76.3}&67.2& 65.2& \underline{82.0} &\multicolumn{3}{c}{--}\\ 
LADDER \cite{goncalves2023learning} &       \multicolumn{4}{c}{\textbf{80.2}}&  \multicolumn{4}{c}{\underline{80.3}}  &  --  & -- &   -- &\multicolumn{3}{c}{--}   \\ 
DE-III (ours)         &    \multicolumn{4}{c}{\underline{79.5}}&       \multicolumn{4}{c}{\textbf{83.7}}&          \textbf{89.3}&          \textbf{88.7}&  \textbf{85.8}  & \multicolumn{3}{c}{\textbf{82.7}}\\
% \textbf{DE-III (-A)}         &   \\
% \textbf{DE-III (-F)}         &    \\
\bottomrule
\end{tabular}
\end{table}

\subsection{Experimental Setup}

\subsubsection{Datasets and Metrics.}

To verify the effectiveness of our proposed approach, we evaluate it on three popular AVER datasets: CREMA-D~\cite{cao2014crema}, MSP-IMPROV~\cite{busso2016msp} and RAVDESS~\cite{livingstone2018ryerson}. CREMA-D consists of 7,442 facial videos with corresponding audio from 96 participants (48 male, 48 female). Each audio-video clip is labeled with one of 6 concrete emotion classes -- anger, disgust, fear, happiness, sadness, and neutrality. RAVDESS consists of 2,880 videos from 24 actors, each enacting eight concrete emotional states. MSP-IMPROV consists of 8,385 audio-video clips from 12 participants (6 male, 6 female) with each clip labeled by both concrete emotional states and continuous emotional states -- valence, arousal and dominance; following previous works~\cite{tsai2019multimodal,chumachenko2022self,gong2022uavm,goncalves2023learning} we use only the continuous labels. We adhered to the protocol in \cite{chumachenko2022self,goncalves2022auxformer}, with 5 separate folds where each fold divides the data into training, validation, and test sets with non-overlapping actor identities. 
We evaluate based on the most commonly-used metrics for each dataset -- F1-Macro and F1-Micro for CREMA-D~\cite{cao2014crema}, Accuracy for RAVDESS~\cite{livingstone2018ryerson} and CCC for MSP-IMPROV~\cite{busso2016msp}.

\subsubsection{Implementation Details.}

All models were trained for up to 20 epochs using early stopping on the validation set, and we report our results on the test set. 
We choose hyper-parameters based on validation set performance.
We use AdamW for optimization with a learning rate of $5 \times 10^{-6}$ and weight decay of $5 \times 10^{-2}$. The face images are extracted from each frame of every video clip and resized to $224\times224$ pixels. We generate optical flow maps using \cite{jaegle2021perceiver} and normalize their magnitude by a standard deviation calculated from the local optical flow magnitude at every pixel position within an entire video clip. We use the pre-trained EfficientNet-B2 from \cite{savchenko2021facial} to extract features from the video frames and optical flow maps. The audio features are extracted using wav2vec2-large-robust \cite{hsu2021robust}. Separate Conformer encoders for video and audio map the extracted features to vectors of 1408-dimension each. Each Conformer block has a hidden dimensionality of 512, with 8 attention heads. The number of blocks in the acoustic, visual, and optical flow Conformers were set to 3, 3, and 2, respectively. For the prediction heads, we use MLPs with hidden dimensionality of 512. Our IFE module (Section~3.3) uses single-head attention \cite{vaswani2017attention} with the linear feed-forward block and the highlighted fusion feature dimensions remain unchanged. Our model was implemented in PyTorch and trained on 2 NVIDIA RTX A5000 GPUs, taking 1 hour.

\subsection{Quantitative Comparison}

In Table~\ref{tab:comparison} we present quantitative results for our method and several existing works:
1) Multi~\cite{tsai2019multimodal}, a transformer-based cross-modal attention fusion method; 2)  MMER \cite{chumachenko2022self}, with multiple self-attention fusion mechanisms; 3) UAVM \cite{gong2022uavm}, a transformer-based feature enhancement model with a shared audio-visual encoder; 4) AuxFormer \cite{goncalves2022auxformer}, a transformer framework with two independent auxiliary branches; 5) LADDER \cite{goncalves2023learning}, a transformer-based cross-attention framework with auxiliary reconstruction tasks. We see that compared with the previous best method LADDER \cite{goncalves2023learning} on CREMA-D, our DE-III achieves higher performance in terms of F1-Micro score, 83.7\% vs. 80.3\%. On MSP-IMPROV, our DE-III attains excellent CCC values of 89.3\% for valence (Val.), 88.7\% for arousal (Aro.), and 85.8\% for dominance (Dom.), establishing a new state-of-the-art for this dataset. Moreover, we also achieve a better accuracy (Acc.) score on RAVDESS compared with the SOTA method, 82.7\% for DE-III vs. 81.6\% for MMER.

\begin{table}[t]
\centering
\fontsize{9.5}{12}\selectfont
\caption{Effectiveness of our inter-modal feature enhancement module (IFE), evaluated on CREMA-D.
}\label{tab:abl_1}
\vspace{6pt}
\begin{tabular}{@{}lcccccccccc@{}}
\toprule
\multirow{2}{*}{Method}&   \multicolumn{2}{c}{Cross attention}&\multicolumn{8}{c}{Accuracy} \\
\cmidrule(r){2-3} \cmidrule{4-11}
 & A-cross& V-cross~~~&\multicolumn{4}{c}{F1-Macro}& \multicolumn{4}{c}{F1-Micro}\\
 \midrule
IFE-Fusion~~      &  \ding{51}&\ding{51}&\multicolumn{4}{c}{77.2}&\multicolumn{4}{c}{82.2}\\ 
IFE-Audio     &  \ding{51}&\ding{55}&\multicolumn{4}{c}{78.3}&\multicolumn{4}{c}{82.2}\\ 
IFE-Video     &  \ding{55}&\ding{51}&\multicolumn{4}{c}{\textbf{79.5}}&\multicolumn{4}{c}{\textbf{83.7}}\\ 
None-IFE&  \ding{55}&\ding{55}&\multicolumn{4}{c}{75.8}&\multicolumn{4}{c}{78.6}\\ 
\bottomrule
\end{tabular}
\end{table}

\subsection{Ablation Studies}

In this section, we evaluate the performance benefit due to various components and design decisions in our model.

\subsubsection{Effects of Inter-modal Feature Enhancement (IFE).}

In the IFE block, we define video attending to audio as V-cross and audio attending to video as A-cross. We first experiment with removing the IFE module (i.e.~without any inter-modality fusion, only RGB images and flow maps, denoted None-IFE). In Table~\ref{tab:abl_1}, we see a large performance drop in this setting -- compared with the best output (from IFE-Video), the F1-Macro and F1-Micro scores decrease by 3.7\% and 5.1\% on the CREMA-D test set, respectively. This suggests that inter-modality fusion plays an important role in improving AVER capabilities. Recall that our model has three prediction heads: IFE-Audio (i.e. using features $\ddot{a}^*$), IFE-Video (i.e. using $\ddot{ov}^*$) and IFE-Fusion (i.e. using their concatenation). While the main results use IFE-Video at inference time, we also report results from the others in Table~\ref{tab:abl_1}.  IFE-Video achieves the best AVER performance, 79.5\% F1-Macro and 83.7\% F1-Micro. The other prediction heads achieve slightly lower though still competitive results.

\subsubsection{Effects of Video Pairwise Attention Enhancement (PAE) Module.}

\begin{table}[t]
\centering
\caption{Effectiveness of different approaches to inter-modal fusion within our model, evaluated on CREMA-D.}
\label{tab:abl_2}
\resizebox{\linewidth}{!}{
\begin{tabular}{@{}cccccccccccc@{}}
\toprule
\multirow{2}{*}{{Model}}  &\multicolumn{2}{c}{Fuse when?} &  \multicolumn{2}{c}{Visual input} &  \multicolumn{2}{c}{Seq. model} & \multicolumn{3}{c}{Fuse how?} &\multicolumn{2}{c}{Accuracy} \\ 
\cmidrule(r){2-3}\cmidrule(){4-5}\cmidrule(lr){6-7}\cmidrule(r){8-10}\cmidrule{11-12}
    &  Early & Late~   & Flow& RGB~ & Conf. & Transf.~ & Concat & Sum & PAE~ & F1-Macro & F1-Micro \\
\midrule
IFE-V-O  & \multicolumn{2}{c}{n/a}     & \ding{51}& & \ding{51}& & \multicolumn{3}{c}{n/a} & 55.4 & 64.9 \\
IFE-V-F   &   \multicolumn{2}{c}{n/a}     & & \ding{51}& \ding{51}& & \multicolumn{3}{c}{n/a} & 76.7  & 81.4 \\
\midrule
IFE-V-FOSC  &   &    \ding{51} & \ding{51}& \ding{51}& \ding{51}& & \ding{51}&& & 78.5 & 81.7 \\
IFE-V-FODC &   & \ding{51} & \ding{51}& \ding{51}& \ding{51}& & \ding{51}&&&77.8 & 82.6 \\
IFE-V-FODS &   &   \ding{51} & \ding{51}& \ding{51}& \ding{51}& & &\ding{51}& & 78.0 & 81.8 \\
\midrule
IFE-V-Early &\ding{51} &      & \ding{51}& \ding{51}& \ding{51}& & &&\ding{51} &{79.2} & {83.0} \\ 
\midrule
{IFE-V-Trans} &   &   \ding{51} & \ding{51}& \ding{51}& & \ding{51}& &&\ding{51} &77.9 & 82.6 \\
\midrule
\textbf{IFE-Video} &   &   \ding{51} & \ding{51}& \ding{51}& \ding{51}& & & & \ding{51} & \textbf{79.5} & \textbf{83.7} \\
\bottomrule
\end{tabular}
}
\end{table}
To demonstrate our ablations on pair-wise attention enhancement (PAE) Module, we categorize different settings as "Fuse when?", "Visual input", "sequential model", and "Fuse how?". Results on CREMA-D are given in Table~\ref{tab:abl_2}, all using the IFE-Video prediction head. We first present results when trained with only one part of the video information, i.e. RGB images only (IFE-V-F), or optical flow maps only (IFE-V-O). We see that IFE-V-O achieves 55.4\% F1-macro and 64.9\% F1-micro. The result shows optical flow information present low capability to distinguish emotions, and it is much weaker than using RGB images only. When combining optical flow maps with RGB images in the full model (IFE-Video), there is a remarkable performance improvement vs.~IFE-V-F. It indicates that the flow maps indeed augment the video feature representations. Next, we replace our PAE with one single conformer followed by one OV-fusion block. To pass the image and optical flow features together into the conformer, we attempt several alternative operations-- temporal concatenation (IFE-V-FOSC), channelwise concatenation (IFE-V-FODC), and summation (IFE-V-FODS). We see (Table~\ref{tab:abl_2}) that our PAE module achieves the highest recognition performance, with 1.0\% improvement over IFE-V-FOSC on F1-macro and 1.1\% improvement over IFE-V-FODC on F1-Micro. These observations indicate that our PAE module is a more effective fusion method for combining visual features and optical flow features. Finally, we explore early fusion and late fusion strategies. We find that by moving OV-fusion block before the Conformer (IFE-V-Early), accuracy decreases slightly vs.~having OV-fusion after the Conformer (IFE-Video), by 0.3\% F1-Macro and 0.7\% F1-Micro. We hypothesise that this is because the additional computation performed beforehand by the Conformer is beneficial in helping the OV-fusion module to determine whether to focus on image or flow information for each time-point. Additionally, we compare our method by replacing the conformer to the vanilla transformer \cite{vaswani2017attention}, the accuracy decreases slightly by 1.6\% and 1.1\%, this demonstrates that the conformer is superior to the vanilla transformer at the image level in capturing changes in facial details from feature representations.

\begin{table}[t]
\centering
\fontsize{9.5}{14}\selectfont
\caption{Effectiveness of different feature extractors and frame-selection strategies for optical-flow, evaluated on CREMA-D for our IFE-Video model variant.}
\label{tab:abl_3}

\begin{tabular}{@{}l@{~~}c@{~~~}c@{~~~}c@{~~}c@{}}
\toprule
\multirow{2}{*}{Feature extractor} & \multirow{2}{*}{Window} & \multirow{2}{*}{Stride} & \multicolumn{2}{c}{Accuracy} \\
 \cmidrule{4-5}
&  & & F1-Macro & F1-Micro \\ 
\midrule
EfficientNet-B2~\cite{savchenko2021facial}  &1 & 1& \textbf{79.5} & \textbf{83.7} \\
\cmidrule(l){2-5}
 &3 & 1& 76.1 & 81.4 \\
 &5 & 1& 74.3 & 80.8 \\
 &7 & 1& 75.2 & 81.6 \\
\cmidrule(l){2-5}
&3 & 3& 77.2 & 82.4 \\
&5 & 5& 76.2 & 80.7 \\
&7 & 7& 78.5 & 82.8 \\
\midrule
DINOv2 \cite{oquab2023dinov2}&1 & 1& 76.8 & 82.6 \\
\bottomrule
\end{tabular}
\end{table}

\subsubsection{Effects of optical-flow extraction variants.}

We next experiment with using different sliding window lengths and strides when extracting the optical flow from the videos. Firstly, we vary the window length while keeping the stride fixed to 1 (i.e.~moving frame by frame). Secondly, we vary both the window length and the stride together (i.e.~non-overlapping windows). The results in Table~\ref{tab:abl_3} show that using a window length of 1 with a stride of 1 performs best. Increased window lengths, with fixed or increasing strides, show consistent drops in performance, with the worst-performing variant having window length of 5 and stride of 1 (achieving 74.3\% F1-Macro, versus 79.5\% for window length and stride of 1). This indicates that temporally-fine-grained information is valuable in increasing the accuracy of emotion recognition. We also experiment with using a different backbone feature extractor for the optical flow, since face images and flow-maps are quite different domains. We choose DINOv2~\cite{oquab2023dinov2}, which has been shown to be robust across many image domains, and fix the window length and stride to 1 (i.e.~the best-performing setting).
However, we find it performs worse than using EffcientNet pre-trained on a large face images dataset, dropping from 79.5\% to 76.8\% F1-Macro and from 83.7\% to 82.6\% F1-Micro.
\begin{figure}[t] 
	\centering
	\includegraphics[width=1\linewidth]{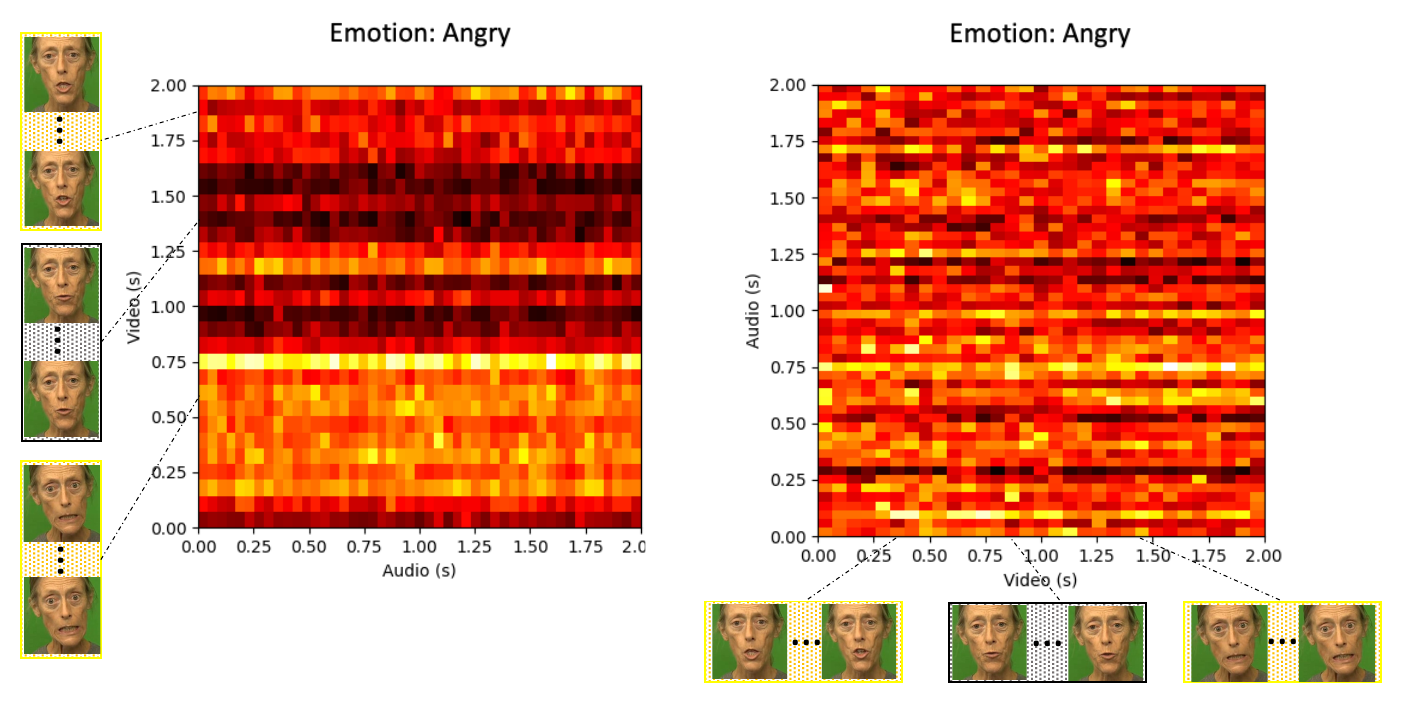}
 
	\caption{Heatmaps showing inter-modality attention weights calculated by IFE-Audio (left) and IFE-Video (right), for an example sequence with emotion `angry'. The horizontal axis corresponds to time-points in one modality, which is fusing in information from the other modality on the vertical axis. Brighter colors indicate stronger attention to the time-point on the vertical axis, from the time-point on the horizontal axis. 
 }
	\label{fig:vis}
\end{figure}

\subsection{Qualitative Analysis}

To better understand the behavior of our model, we visualize the inter-modal fusion weights $\alpha_{ij}$ for IFE-Audio and IFE-Video (see Section~3.3) in Figure~\ref{fig:vis}. The brightness of each location in the heatmap represents the strength with which the modality on the horizontal axis is attending to that on the vertical axis, at that particular time-point. The pattern of attention varies considerably for different points along the horizontal axis, showing that the model does not attend to fixed, specific points in the other modality, but adapts depending on the current features, and presumably the varying emotional states depicted in the video.
Notably, the heatmaps do not exhibit a bright diagonal line; this indicates that time-points generally attend not to the corresponding time-point in the other modality, but to other (presumably relevant or informative) time-points. Overall these results suggest that our inter-modal feature enhancement module can selectively fuse the useful information from each modality into the other.

\section{Conclusion}

We have presented a new model, DE-III, for audio-visual emotion recognition, which combines intra- and inter-model feature enhancement in a unified framework. DE-III introduces a pair-wise attention fusion method that integrates explicit facial detail changes between video frames, captured by optical flow. It not only improves the distinguishability of features within each visual modality, but also further increases the effectiveness of subsequent inter-modal feature interactions. Our results demonstrate that DE-III enhances emotion recognition by optimally fusing the information available in different modalities. Indeed, our model achieves state-of-the-art performance on three popular datasets, for both concrete and continuous emotion labels.
\bibliographystyle{splncs04}
\bibliography{icme2023template}
\end{document}